\documentclass{article} 
\usepackage{iclr2021_conference,times}

\usepackage{amsmath,amsfonts,bm}

\def\eqref#1{equation~\ref{#1}}

\def\1{\bm{1}}

\def\rvc{{\mathbf{c}}}

\def\rvx{{\mathbf{x}}}

\def\rvz{{\mathbf{z}}}

\DeclareMathAlphabet{\mathsfit}{\encodingdefault}{\sfdefault}{m}{sl}
\SetMathAlphabet{\mathsfit}{bold}{\encodingdefault}{\sfdefault}{bx}{n}

\DeclareMathOperator*{\argmin}{arg\,min}

\usepackage{hyperref}
\usepackage{url}
\usepackage{graphicx}
\usepackage{longtable}
\usepackage{booktabs}
\usepackage{placeins}
\usepackage{svg}
\usepackage{multirow}

\usepackage[group-separator={,}]{siunitx}
\usepackage{subfigure}

\usepackage{amsmath,amssymb,amsfonts}
\usepackage{xcolor}

\newcommand{\drawback}[1]{\textcolor{red}{#1}}

\title{Is Disentanglement all you need? Comparing Concept-based \& Disentanglement Approaches}

\author{Dmitry Kazhdan\textsuperscript{a}, Botty Dimanov\textsuperscript{a}, Helena Andres Terre\textsuperscript{a}, \\ 
\textbf{Mateja Jamnik\textsuperscript{a}, Pietro Li\`{o}\textsuperscript{a}, Adrian Weller\textsuperscript{a, b}} \\
\textsuperscript{a}The University of Cambridge, Cambridge, UK\\
\textsuperscript{b}The Alan Turing Institute, London, UK \\
\texttt{\{dk525, ha376\}@cam.ac.uk} \\
}

\iclrfinalcopy 
\begin{document}

\maketitle



\begin{abstract}

Concept-based explanations have emerged as a popular way of extracting human-interpretable representations from deep discriminative models. At the same time, the disentanglement learning literature has focused on extracting similar representations in an unsupervised or weakly-supervised way, using deep generative models. Despite the overlapping goals and potential synergies, to our knowledge, there has not yet been a systematic comparison of the limitations and trade-offs between concept-based explanations and disentanglement approaches. In this paper, we give an overview of these fields, comparing and contrasting their properties and behaviours on a diverse set of tasks, and highlighting their potential strengths and limitations. In particular, we demonstrate that state-of-the-art approaches from both classes can be data inefficient, sensitive to the specific nature of the classification/regression task, or sensitive to the employed concept representation.

\end{abstract}


\section{Introduction} \label{introduction}

Recent years have witnessed a dramatic increase in work on Explainable Artificial Intelligence (XAI) approaches, exploring ways of improving the explainability of deep learning models, and making them more transparent and trustworthy~\citep{guidotti2018survey,carvalho2019machine,adadi2018peeking}. A novel strand of XAI approaches relies on \textit{concept-based} explanations, presenting explanations in terms of higher-level, human-understandable units (e.g., a prediction for an image of a car may rely on the presence of wheel and door concepts)~\citep{tcav,kazhdan2020now,koh2020concept,chen2020concept}. 
At the same time, recent work on \textit{disentanglement learning} aims to learn input data representations that have useful properties, such as better predictive performance~\citep{locatello2019challenging},  fairness~\citep{creager2019flexibly}, and interpretability~\citep{adel2018discovering}. These approaches assume that high-dimensional real-world data (e.g., images) is generated from a relatively-smaller number of \textit{latent factors} of variation~\citep{bengio2013representation}. 

Despite their overlapping goals and potential synergies, these two fields are typically not considered together. In this paper, we give an overview of these two fields and argue that concepts are a type of interpretable factors of variations. We conduct a systematic comparison between disentanglement learning and concept-based explanation approaches, showcasing their limitations and trade-offs to underscore their underlying assumptions and causes of failures due to training, labelling, and choice of objective functions.


%
We present a comprehensive \textit{library}\footnote{https://github.com/dmitrykazhdan/concept-based-xai}, implementing a range of state-of-the-art concept-based, and disentanglement learning approaches. Using this library, we compare approaches across a variety of tasks and datasets. In particular, we demonstrate that state-of-the-art approaches can be data inefficient, sensitive to the specific nature of the end task, or sensitive to the nature of the concept values. 
All experimental setups will be released as part of the library, and will serve as a foundation for future research aiming to address the limitations of these approaches.



\section{State of the Art: Analysis and Limitations}
\label{sec:methodology}

\textbf{Concept Bottleneck Models (CBMs)}: CBMs~\citep{koh2020concept,chen2020concept} assume that the training data consists of inputs $\rvx_i$, task labels $y_i$, and additional \textit{concept annotations} $\rvc_i$: $(\rvx_{1}, \mathbf{c}_{1}, y_{1}), ... (\rvx_{N}, \mathbf{c}_{N}, y_{N})$. CBMs are models which are interpretable by design, separating input processing into two distinct steps: (i) predicting concepts from the input, and (ii) predicting task labels from the concepts. These can be represented as a concept predictor function $g_{cp}$, and a label predictor function $f_{lp}$, such that $f_{lp}(g_{cp}(\rvx)): \mathcal{X} \to \mathcal{C} \to \mathcal{Y}$, where $\mathcal{X}, \mathcal{C}, \mathcal{Y}$ denote the input, concept, and output representations, respectively. Further details can be found in Appendix~\ref{sec:appendix-cbm}.

\drawback{\textbf{Limitations}}: (i) CBMs typically require a large number of concept annotations during training, making them data inefficient. 
(ii) If the downstream task is dependent on information not contained in the provided concepts, then CBMs are either incapable of accurately predicting task labels, or their concept representations contain information ``impurities'' that are not related to the concepts (see Appendix~\ref{sec:appendix-cbm} for further details). The former hinders performance, while the latter hinders interpretability, as it becomes unclear whether a model is using concept or non-concept information.

\textbf{Post-hoc Concept Extraction (CME)}: CME approaches~\citep{kazhdan2020now,yeh2019completeness, tcav} extract concepts from pre-trained models. Given a trained model $f(\rvx) = y$, CME relies on $f$ to obtain a feature extractor $\phi(\rvx) : \mathcal{X} \to \mathcal{H}$, computing a higher-level representation of the input. Using this feature extractor, CME extracts models $f_{lp}(g_{cp}(\phi(\rvx))) : \mathcal{X} \to \mathcal{H} \to \mathcal{C} \to \mathcal{Y}$. Similarly to CBMs, these models compute a concept representation using $g_{cp}$, and then use it to compute the task labels via $f_{lp}$. Importantly, instead of using the raw input $\rvx$, these methods rely on a reduced representation returned by $\phi$, typically making them more data-efficient. Further details can be found in Appendix \ref{sec:appendix-cme}.

\drawback{\textbf{Limitations}}: (i) CME approaches assume that concepts can be reliably predicted from the hidden space of a trained model (i.e., that the representation computed by $\phi$ is informative of the concepts). Thus, CME requires a powerful pre-trained model in order to extract concepts reliably. (ii) Existing CME approaches either extract concepts in a semi-supervised fashion~\citep{kazhdan2020now}, which requires the concepts to be known beforehand, or in an unsupervised fashion~\citep{yeh2019completeness}, which requires the extracted concepts to be manually inspected and annotated afterwards.

\textbf{Unsupervised Disentanglement Learning:} Disentanglement learning approaches, such as VAEs~\citep{vae,higgins2016beta}, assume that input data $\rvx$ is generated by a set of causal latent factors $\rvz$, sampled from a distribution $p(\rvz)$, such that the input $\rvx$ is sampled from $p(\rvx|\rvz)$. Consequently, they aim to \textit{automatically} learn mappings between the input data $\rvx$ and lower-dimensional space $\rvz$, that describes the corresponding factors of variations.


\drawback{\textbf{Limitations}}: Recently, \citet{locatello2019challenging} showed that unsupervised VAEs can learn infinitely many different possible latent factor solutions that are no guaranteed to learn factors with desirable properties, such as human-interpretability, fairness, or high downstream task performance, without explicit supervision.


\textbf{Weakly-supervised Disentanglement Learning}: Weak supervision can be used to induce desirable properties on the learned factors, where a (small) set of data-points is labelled with ground-truth latent factor annotations~\citep{locatello2020weakly}.
Importantly, knowledge of the ground-truth data generative process is not available in most practical scenarios.
Thus, a human annotator usually relies on domain knowledge of the task to provide labels of his/her \textit{perception} of the factors of variation, instead of the ground-truth latent generative factors (i.e., factors of variation used by the environment).
Consequently, we argue that these labels correspond to \textit{concepts} (i.e., human perceived factors of variation dictated by domain knowledge). Hence, the latent factors $\mathbf{z}$ used in weakly-supervised disentanglement learning can be seen as concepts $\mathbf{c}$. Further details can be found in Appendix \ref{sec:appendix-wvae}.




\drawback{\textbf{Limitations}}: (i) The weak supervision requirement of WVAE approaches imply that a user must have some knowledge of the underlying latent factors. (ii) VAE-based approaches predominantly rely on a combination of reconstruction loss and regularisation loss (the latter encourages disentanglement of the latent space). 
This implies that WVAEs may exhibit \textit{bias}, since concepts having a larger variability in the input space (which we refer to as \textit{loud} concepts) will have a relatively larger effect on the reconstruction loss, and will therefore be learned first. This has important fairness implications, particularly with respect concepts such as colour, as will be explored later on.



\section{Experiments} \label{experiments}


\textbf{Datasets} We evaluate the compared methods on datasets commonly-used for benchmarking disentanglement learning approaches~\citep{locatello2020weakly}, in which the underlying generative factors describe all possible data variations: \textit{dSprites}~\citep{dsprites17}, and \textit{shapes3D}~\citep{3dshapes18}. These datasets allow fine-grained control over the utilised concept representations and higher-level tasks, and have been previously used in work on concept explanations, including~\citep{kazhdan2020now}. Further details regarding the datasets can be found in Appendix~\ref{appendix-datasets}. 

\textbf{Benchmarks} We compare and evaluate the following state-of-the-art approaches described in Section~\ref{sec:methodology}: multi-task CBMs~\citep{koh2020concept}, CME with gradient boosted tree (GBT)~\citep{hastie2009elements} concept extraction functions~\citep{kazhdan2020now}, unsupervised VAEs, and Adaptive-ML-VAE (WVAEs)~\cite{locatello2020weakly}, for which weak supervision is provided by feeding in pairs of data samples, differing in only a few concepts at a time. While CME and CBM predict concepts explicitly, in the case of WVAEs we train a GBT, as in \citep{locatello2020weakly}, to predict concept values from the latent representation.  
Further details regarding these benchmarks can be found in Appendix~\ref{sec:appendix-benchmarks}.



\section{Results \& Discussion}


\paragraph{Data Efficiency}  As discussed in Section \ref{sec:methodology}, CBMs assume that all input data-points are explicitly labelled with concept annotations (i.e. have the associated concept labels for every data point), while WVAEs assume this information is provided implicitly via pairs of points. In practice, obtaining concept annotations may be expensive or infeasible, especially for tasks with a large number of concepts. Consequently, it is important to consider the amount of annotations required by these approaches for accurate concept extraction (we refer to this as \textit{data efficiency}).

\begin{figure}[t]
    \centering
    \includegraphics[height=3.5cm]{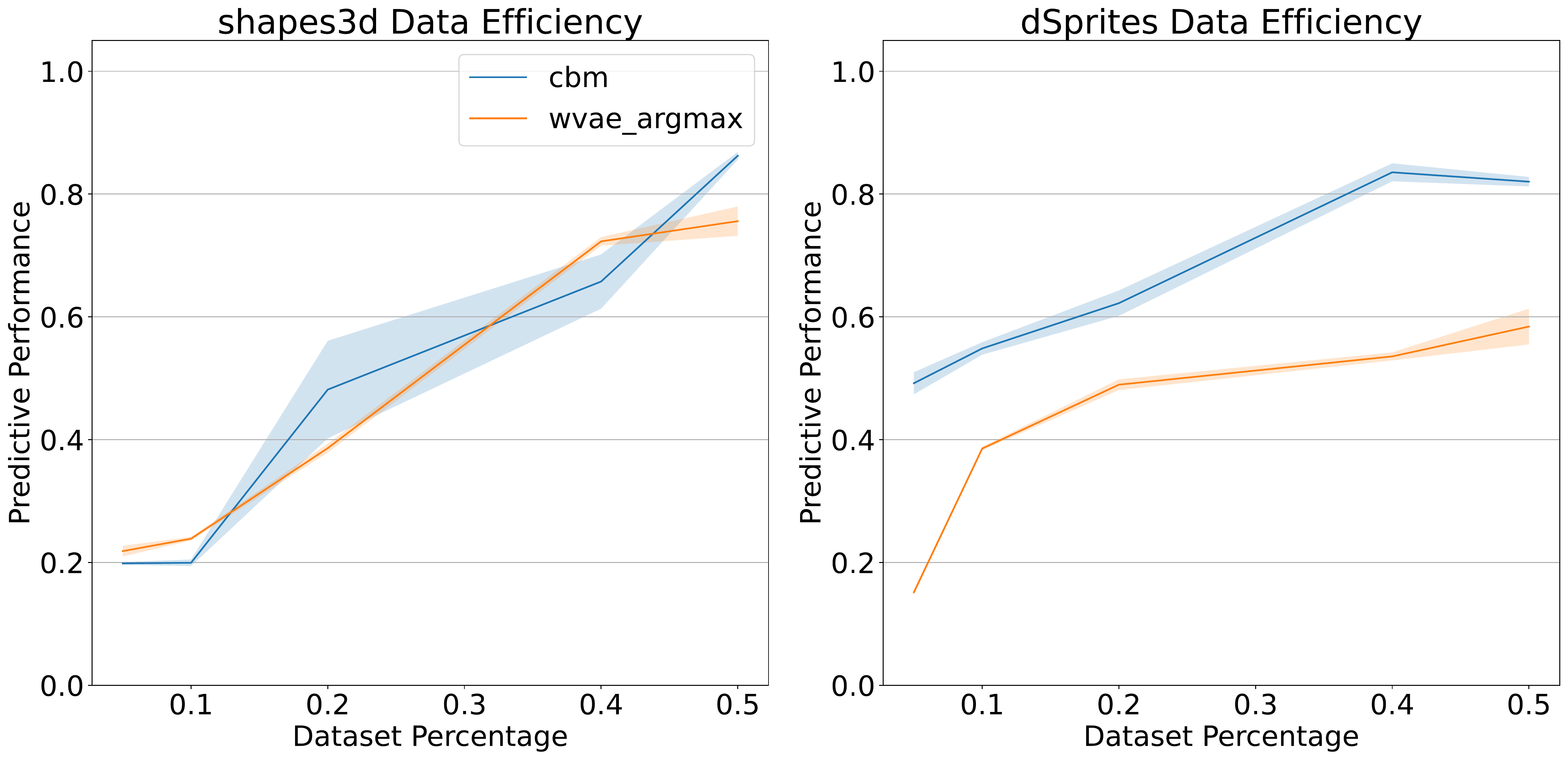} 
    \caption{\normalsize{Average concept predictive accuracy (y-axis) of WVAE and CBM for shapes3D (left) and dSprites (right) across a progressively larger number of available concept annotations as a percentage of the total number of concept annotations (x-axis).}}
    \label{fig:data_efficiency}
\end{figure}

In this experiment we compare the amount of supervision necessary to extract concept representation for: (1) CBM and (2) WVAE. We train both models with progressively smaller labelled datasets containing labels of \textit{all} concepts, observing the corresponding change in concept predictive accuracy, averaged over all concepts. Figure~\ref{fig:data_efficiency} demonstrates that the concept predictive accuracy for both approaches decreases significantly with the labelled dataset sizes, and that both CBMs and WVAEs require almost 40\% of all training points to be labelled to achieve above 60\% average predictive accuracy. Overall, CBMs are more data efficient than WVAEs, which is likely due to the fact that they are explicitly trained to optimise for concept predictive accuracy. Importantly, unsupervised VAEs do not rely on concept annotations at all, and CME is highly sensitive to the end task (as will be shown below). Thus, these approaches are not included in this setup.

\begin{figure*}[t]
    \centering
    \includegraphics[width=\textwidth]{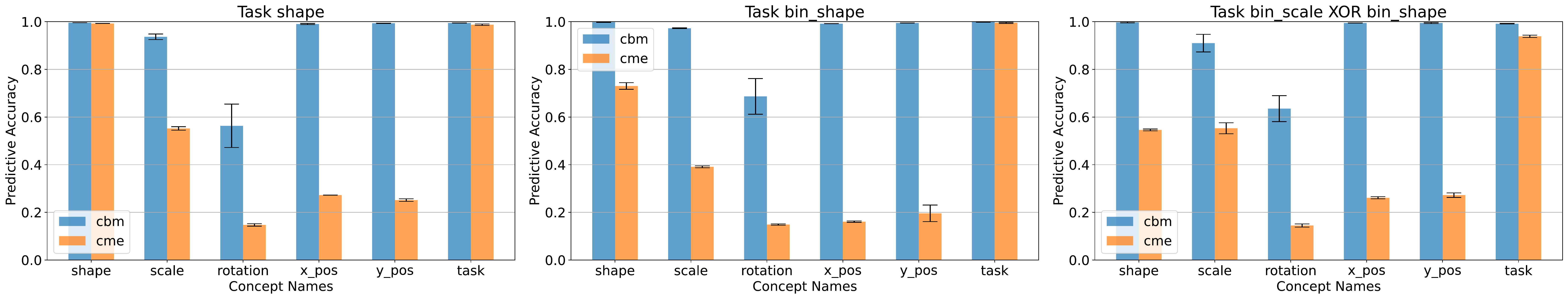}
    \caption{\normalsize{Comparison between the per-concept and downstream task predictive accuracy for CME and CBM across tasks of varying dependence on the concept values: full dependence on one concept (left), partial dependence on one concept (center), partial dependence on two concepts (right). For every subplot, the rightmost column (\text{Task}) represents the predictive performance for the task labels.}}
    \label{fig:task_variance}
\end{figure*}

\paragraph{Concept-to-Task Dependence}  In contrast to both CBMs and WVAEs, CME is explicitly designed to be data efficient. However, CME relies on the hidden representation quality of its corresponding pre-trained model, as discussed in Section \ref{sec:methodology}. Here, we compare the concept predictive accuracy of CME across tasks with a varying dependence on the provided concepts, as shown in Figure~\ref{fig:task_variance}. In all cases, the CME models were trained with merely 500 labelled datapoints (approximately 1\% of the concept annotations). We also provide results for CBM models trained on the entire datasets of concept annotations (which defines an upper bound for the concept predictive accuracy). Further details can be found in Appendix~\ref{sec:appendix-task-variation-exp} and Appendix~\ref{sec:appendix-task-variation-figs}.

Firstly, Figure~\ref{fig:task_variance} demonstrates that CME successfully exracted the concept information necessary to achieve high predictive accuracy on the end task (indicated by the high `Task' performance on all three tasks). However, concept predictive accuracy progressively declined as the task labels had progressively less reliance on the specific concept values. Thus, the CME predictive performance of each concept is highly dependant on how much this concept affects the end task label.



\begin{figure*}[t]
    \centering
    \includegraphics[scale=0.09]{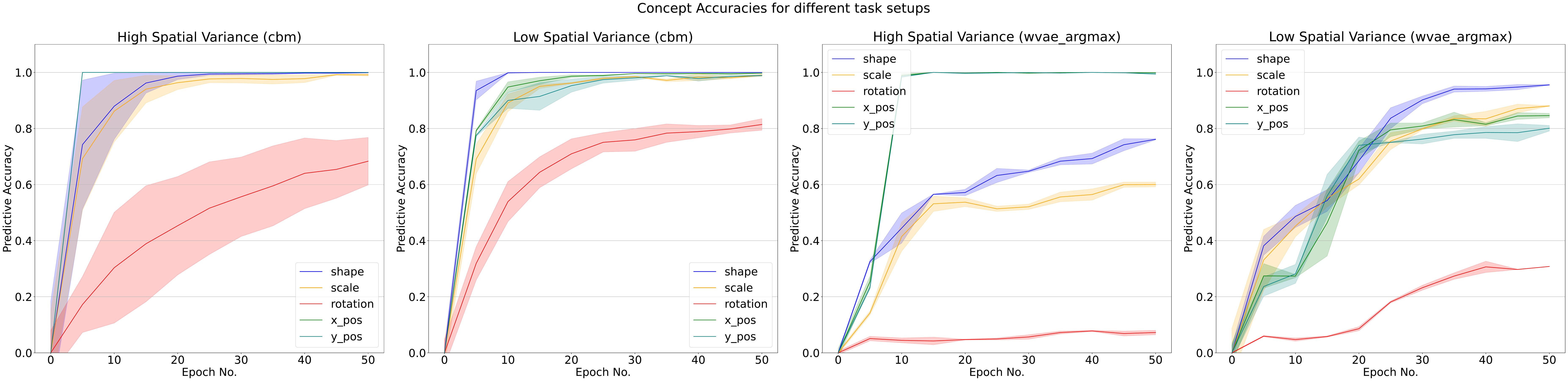}
    \caption{\normalsize{Comparison between different concept loudness setups. `High Spatial Variance' refers to a dataset having $x\_pos$ and $y\_pos$ concepts significantly louder that the rest, whereas `Low Spatial Variance' refers to a dataset where these concepts have similar loudness to the other concepts. The subfigures on the left show the performance of CBM, whilst subfigures on the right show performance of WVAE. Importantly, the other 3 concepts are significantly affected by the loudness of $x\_pos$ and $y\_pos$ in the case of WVAE.}}
    \label{fig:concept_variance}
\end{figure*}

\paragraph{Concept Variance Fragility} As discussed in Section \ref{sec:methodology}, WVAE approaches are implicitly biased toward concepts whose values have a higher variance in the input space (the \textit{louder concepts}). Consequently, we hypothesise that it is harder and significantly slower for WVAEs to learn the \textit{quieter} concepts. We compare the effect of \textit{concept loudness} on the ability of WVAEs and CBMs to learn concept representations, by inspecting how the predictive accuracy for every concept changes over time as the model is trained. Figure~\ref{fig:concept_variance} illustrates that relative ``concept loudness'' can have a significant effect on the WVAEs ability to learn concepts (i.e., the performance of quieter concepts degrades significantly). On the other hand, CBMs are significantly less affected by this phenomenon. Further details regarding this experimental setup can be found in Appendix \ref{sec:appendix-c-var-exp}. Importantly, colour combinations can have an effect on concept loudness. Hence, a concept representing colour (e.g., the colour of an object) can vary in loudness, depending on the specific colour combinations represented by that concept. This is explored further in Appendix \ref{sec:appendix-c-var-exp} and Appendix \ref{sec:appendix-c-var-figs}.

\textbf{Conclusions} We analyse existing concept-based and disentanglement learning approaches with respect to three important properties: \textit{data efficiency}, \textit{concept-to-task dependence}, and \textit{concept loudness}. We demonstrate that WVAE and CBM approaches are data inefficient, whilst CME is sensitive to the dependence between the concepts and the end task. Additionally, WVAE can be sensitive to the relative loudness of concepts. Overall, we believe these findings can be used for selecting the most suitable concept extraction approach, as well as for highlighting important shortcomings of state-of-the-art methods, that should be addressed by future work.


\section*{Acknowledgements}

AW acknowledges support from a Turing AI Fellowship under grant EP/V025379/1, The Alan Turing Institute under EPSRC grant EP/N510129/1 and TU/B/000074, and the Leverhulme Trust via CFI.

\bibliography{references}
\bibliographystyle{iclr2021_conference}

\appendix

\section{Implementation}

\subsection{Datasets} \label{appendix-datasets}

\subsubsection{dSprites}
dSprites~\citep{dsprites17} consists of 2D $64 \times 64$ pixel black-and-white shape images, procedurally generated from all possible combinations of 6 ground truth independent concepts (color, shape, scale, rotation, x and y position) for a total of $1 \times 3 \times 6 \times 40 \times 32 \times 32 = 737280$ total images. Table~\ref{dsprites_features} lists the concepts and their corresponding values, while Figure~\ref{fig:dataset-img} (a) presents a few examples. 

\begin{table}[h!]
    \centering
    \caption{dSprites concepts and values}
    \label{dsprites_features}
    \begin{tabular}{cc}
    \\
    \toprule
     \textbf{Concept name} & \textbf{Values} \\ 
     \bottomrule
Color               & white                                         \\ 
Shape               & square, ellipse, heart                        \\ 
Scale               & 6 values linearly spaced in $ [0.5, 1] $      \\ 
Rotation            & 40 values in $[0, 2 \pi]  $                    \\ 
Position X          & 32 values in $[0, 1]$                         \\ 
Position Y          & 32 values in $[0, 1]$                         \\ 

     \bottomrule \\
    \end{tabular}
\end{table}


\subsubsection{3D Shapes}
\begin{figure}[h!]
\setlength\tabcolsep{0pt}
\centering
\begin{tabular}{cc}
      \includegraphics[width=.33\linewidth]{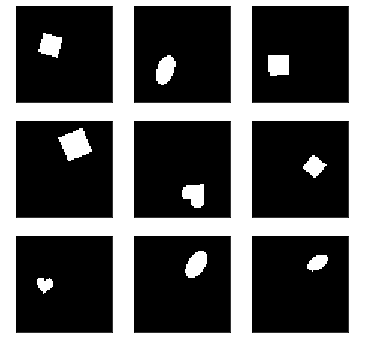} &   \includegraphics[width=.33\linewidth]{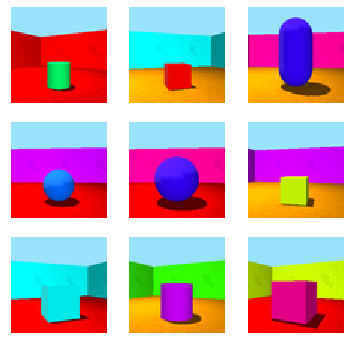} \\
    {\scriptsize (a) dSprites \par} & {\scriptsize (b) 3dshapes \par} \\[6pt]
    \end{tabular}
\caption{Sample images from dSprites (a) and 3dshapes (b).}
\label{fig:dataset-img}
\end{figure}

The 3dshapes~\citep{3dshapes18} dataset comprises of $64 \times 64$ pixel coloured 3D shape images, generated from all possible combinations of six latent factors -- floor hue, wall hue, object hue, scale, shape, and orientation. Table~\ref{tbl:3dshapes} lists the concepts, and corresponding values, while Figure~\ref{fig:dataset-img} (b) presents some examples. 

\begin{table}[h!]
    \centering
    \caption{3dshapes concepts and values}
    \label{tbl:3dshapes}
    \begin{tabular}{cc}
    \\
    \toprule
     \textbf{Concept name} & \textbf{Values} \\ 
     \bottomrule
     Floor hue & 10 values linearly spaced in $[0, 1]$ \\
     Wall hue & 10 values linearly spaced in $[0, 1]$ \\
     Object hue & 10 values linearly spaced in $[0, 1]$  \\
     Scale & 8 values linearly spaced in $[0, 1]$ \\
     Shape & 4 values in $[0, 1, 2, 3]$ \\
     Orientation & 15 values linearly spaced in $[-30, 30]$ \\
     \bottomrule \\
    \end{tabular}
\end{table}

\subsection{Benchmarks}
\label{sec:appendix-benchmarks}
\subsubsection{CBMs} \label{sec:appendix-cbm}

As discussed in Section~\ref{sec:methodology}, CBM-based approaches train models that process input data-points in two distinct steps: (i) predicting concepts from the input, and (ii) predicting the task labels from the concepts, which are represented as a concept predictor function $g_{cp}$, and a label predictor function $f_{lp}$, such that $f_{lp}(g_{cp}(\mathbf{x})) : \mathcal{X} \to \mathcal{C} \to \mathcal{Y}$, where $\mathcal{X}, \mathcal{C}, \mathcal{Y}$ denote the input, concept, and output spaces respectively.

\citet{koh2020concept} introduce several different training regimes for learning $f_{lp}$ and $g_{cp}$ from the input data, concept, and task labels: 

\textbf{Independent}: $\hat{g}_{ind} = \argmin_{g} \sum_{i} \mathcal{L}_{c}(g(\rvx_{i}), \mathbf{c}_{i})$, and $\hat{f}_{ind} = \argmin_{f} \sum_{i} \mathcal{L}_{y}(f(\mathbf{c}_{i}), y_{i})$

\textbf{Sequential}: $\hat{g}_{seq} = \argmin_{g} \sum_{i} \mathcal{L}_{c}(g(\rvx_{i}), \mathbf{c}_{i})$, and $\hat{f}_{seq} = \argmin_{f} \sum_{i} \mathcal{L}_{y}(f(\hat{g}_{seq}(\rvx_{i})), y_{i})$

\textbf{Joint}: $\hat{g}_{joint}, \hat{f}_{joint} = \argmin_{f, g} \sum_{i} \mathcal{L}_{y}(f(g(\rvx_{i})), y_{i}) + \lambda \mathcal{L}_{c}(g(\rvx_{i}), \mathbf{c}_{i})$. In this case the $\lambda$ parameter trades off the importance of concept information contained in $\mathbf{c}$, with other information contained in the data which is relevant for predicting the labels $y$. 

In the above, $\mathcal{L}_{c}$ and $\mathcal{L}_{y}$ refer to the concept prediction loss and task label loss, respectively. Further details can be found in \citet{koh2020concept}. 

Importantly, the Independent and Sequential regimes require the task labels to be highly-dependent on the concept information, as it is the only information available to the label predictor. The Joint regime alleviates this issue, by allowing non-concept information to pass through the bottleneck as well. Whilst this modification can improve predictive performance of a CBM, it implies that concept and non-concept information are both mixed in the bottleneck of the CBM. This harms interpretability, as it becomes unclear how a CBM uses concept and non-concept information when producing task labels.

\textbf{Multi-task}: In the original CBM work~\citep{koh2020concept}, all concepts are assumed to be binary. Consequently, the concept predictor is implemented as a single-output model with a sigmoid output layer, having $|neurons| = |concepts|$ (i.e., one neuron per concept). 

However, many other datasets (including those used in this work) contain multi-valued concepts. Consequently, we propose a new type of bottleneck model, namely a \textit{multi-task CBM} model, where the bottleneck layer consists of several output heads, and each concept $c_i$ is learned in a separate output head. Each output head consists of $n_i$ softmax units, where $n_i$ represents the number of unique concept values for concept $c_i$. For example, dSprites has 6 concepts; hence, our multi-task  model will have 6 different outputs, each with an individual sparse-categorical cross-entropy loss. 

Importantly, the concept outputs are all concatenated and fed together into the concept-to-output model. Thus, similarly to the original CBM, the concepts are trained jointly. However, in contrast to the classical CBM setting, this set-up ensures that concept values of the same concept are mutually-exclusive (i.e., an image cannot be both a square and a heart), since they are now separated across multiple outputs. We found that this setup gave better performance than the standard CBM, for datasets with multiple multi-valued concepts (including dSprites and shapes3d).

\subsubsection{CME} \label{sec:appendix-cme}

CME trains concept predictor and label predictor models in a semi-supervised fashion, by relying on the hidden space of a pre-trained model. Given a pre-trained model $f(\mathbf{x}) = y : \mathcal{X} \to \mathcal{Y}$ predicting task labels $y$ from input $\mathbf{x}$, CME uses the hidden space $\mathcal{H}$ of $f$ to train a concept predictor model $\hat{g}_{cp}(\mathbf{h}) = \mathbf{c} : \mathcal{H} \to \mathcal{C}$, using a provided set of concept annotations $\mathbf{c}$ (which is assumed to be small). The optimisation problem is defined as $\hat{g}_{cp} = \argmin_{g} \sum_{i} \mathcal{L}_{c}(g(\phi(\rvx_{i})), \mathbf{c}_{i})$, where $\phi : \mathcal{X} \to \mathcal{H}$ is assumed to be a feature extractor function, returning activations of $f$ at a given layer. Next, CME trains a task labeller function $\hat{f}_{lp}(\mathbf{c}) = y : \mathcal{C} \to \mathcal{Y}$, which is trained using $\hat{f}_{lp} = \argmin_{f} \sum_{i} \mathcal{L}_{y}(f(\hat{g}_{cp}(\phi(\rvx_{i}))), y_{i})$. 

In the above, $\mathcal{L}_{c}$ and $\mathcal{L}_{y}$ refer to the concept prediction loss and task label loss, respectively. Further details, including the choice of layer(s) to use for $\phi$,  can be found in \citet{kazhdan2020now}.

\subsubsection{Weakly-supervised Disentanglement Learning} \label{sec:appendix-wvae}

The standard optimization objective adopted by many existing disentanglement learning approaches is shown in Equation \ref{eq:uvae}. Here, the aim is to jointly learn an encoder model (representing $q_{\phi}(\mathbf{z} | \mathbf{x})$), and a decoder model (representing $p_{\theta}(\mathbf{x} | \mathbf{z})$), which model the underlying data generative process.

\begin{equation}   \label{eq:uvae} 
\max_{\phi, \theta} \mathbb{E}_{p(\rvx)}[\mathbb{E}_{q_{\phi}(\rvz|\rvx)}[\log  p_{\theta}(\rvx | \rvz)] - D_{KL}(q_{\phi}(\rvz|\rvx) || p(\rvz))]
\end{equation} 

As discussed in Section \ref{sec:methodology}, work in \citet{locatello2019challenging} showed that without supervision, unsupervised VAE-based approaches can learn an infinite amount of possible solutions, guided by implicit biases that are not explicitly controlled for. Consequently, a range of recent work explored methods to semi/weakly-supervised disentanglement learning.

In particular, work in \citet{locatello2020weakly} assumes that input data is provided as pairs of points $(\rvx_{1}, \rvx_{2})$ during training, with every pair of points differing by only a few factors of variation at a time. Consequently, the optimization objective is modified as shown in Equation \ref{eq:wvae}. 

\begin{equation} \label{eq:wvae}
\begin{gathered}
\max_{\phi, \theta} \mathbb{E}_{(\rvx_1, \rvx_2)}\mathbb{E}_{\hat{q}_{\phi}(\mathbf{\hat{z}}|\rvx_1)}\log  p_{\theta}(\rvx_1 | \mathbf{\hat{z}}) + \mathbb{E}_{\hat{q}_{\phi}(\mathbf{\hat{z}}|\rvx_2)}\log  p_{\theta}(\rvx_2 | \mathbf{\hat{z}}) \\ 
- \beta D_{KL}(\hat{q}_{\phi}(\mathbf{\hat{z}}|\rvx_1) || p(\mathbf{\hat{z}})) - \beta D_{KL}(\hat{q}_{\phi}(\mathbf{\hat{z}}|\rvx_2) || p(\mathbf{\hat{z}}))
\end{gathered}
\end{equation}

In the above, $\hat{q}$ refers to a modified version of the approximate posterior $q$, defined as show in Equation \ref{eq:wvae_post}, and $a$ refers to an averaging function.

\begin{equation} \label{eq:wvae_post}
\begin{gathered}
\hat{q}_{\phi}(\hat{z}_i | \rvx_1) =  a(\hat{q}_{\phi}(\hat{z}_i | \rvx_1), \hat{q}_{\phi}(\hat{z}_i | \rvx_2)),  \ \ \forall{i \in \hat{S}}  \\
\hat{q}_{\phi}(\hat{z}_i | \rvx_1) = \hat{q}_{\phi}(\hat{z}_i | \rvx_1), \ \ \text{else}
\end{gathered}
\end{equation}

The update rule for $\hat{q}_{\phi}(\hat{z}_i | \rvx_2)$ is obtained in an analogous manner. In the above, set $\hat{S}$ is the set of chosen latent factors, which is obtained by selecting latent factors with smallest $D_{KL}(\hat{q}_{\phi}(\hat{z}_i | \rvx_1) || \hat{q}_{\phi}(\hat{z}_i | \rvx_2))$. Further details can be found in \citet{locatello2020weakly}.

\subsection{Experimental Details}

\subsubsection{Concept-to-Task Dependence} \label{sec:appendix-task-variation-exp}
\textbf{dSprites} 
For dSprites, we use a reduced version of the dataset introduced in \citet{kazhdan2020now}, namely: we select $16$ of the $32$ values for \textit{Position X} and \textit{Position Y} (keeping every other value only), and select $8$ of the $40$ values for \textit{Rotation} (retaining every 5th value) to decrease the $737280$ samples to $3*6*8*16*16=36864$ samples. 

We define 3 separate tasks: \textit{shape}, \textit{bin\_shape}, and \textit{bin\_scale XOR bin\_shape}. The \textit{shape} task consists of predicting the value of the shape concept (i.e. $y=c_{shape}$). The \textit{bin\_shape} task consists of predicting whether the shape concept is a heart or not (i.e., the concept is now binarized into two values, instead of three: $y=((c_{shape}==0) \ OR \ (c_{shape}==1))$). The \textit{bin\_scale XOR bin\_shape} task is defined as: $y = ((c_{shape} == 2) \  XOR \ (c_{scale} > 2))$

\textbf{shapes3d} For shapes3d, we relied on a reduced version of the dataset: we select every other value from all concepts, except \textit{shape}, for which we keep all values, for a total of:  $5 * 5 * 5 * 4 * 4 * 7 = 14000$ samples

Similarly to dSprites, we define 3 tasks: The \textit{shape} task consists of predicting the value of the shape concept (i.e. $y=c_{shape}$). The \textit{bin\_shape} task is defined as $y = (c_{shape} >= 2)$ (i.e., whether it's the first two values, or the last two values). The task \textit{bin\_scale XOR bin\_shape} task is defined as: $y = ((c_{shape} >= 2) \ XOR \ (c_{scale} > 2))$.

For both datasets, the 3 tasks represent a gradual progression from a task which loses no information about one of the concepts (Task 1 is equivalent to the \textit{shape} concept), to a task which loses information about all concepts (Task 3 only loosely depends on the exact \textit{shape} and \textit{scale} concept values).

\subsubsection{Data Efficiency}

For the data efficiency experiments, we rely on the same dataset setups as in the Concept-to-Task Dependence experiments, for both dSprites and shapes3d.

\subsubsection{Concept Variance Fragility} \label{sec:appendix-c-var-exp}

\textbf{dSprites} We define two tasks for the spatial variance experiment. For the first one, we sub-select the following concept values: $shape \in [0, ..., 3], \ scale \in [0, ..., 5], \ range \in [0, ..., 19], \ x\_pos \in [0, 10, 20, 30], \ y\_pos \in [0, 10, 20, 30]$. For the second task, we sub-select the following concept values: $shape \in [0, ..., 3], \ scale \in [0, ..., 5], \ range \in [0, ..., 19], \ x\_pos \in [0, ..., 3], \ y\_pos \in [0, ..., 3]$. Importantly, both tasks consist of exactly the same concepts and have the same number of values per concept. However, the input variance of the first task wrt. the $x\_pos$ and $y\_pos$ concepts is significantly larger than for the second task (i.e., $x\_pos$ and $y\_pos$ are relatively louder in the first setup), since the difference in the input space between the different concept values is now larger.

For the colour variance experiment, we introduce an extra \textit{colour} concept to dSprites, which can take on two values. Consequently, we define three tasks, with the colour concept taking on different colour combinations. For all three tasks, we set the first value to be the colour \textit{green} (RGB $[0, 1, 0]$). Next, we vary the value of the second colour to be \textit{purple} (RGB $[1, 0, 1]$), \textit{blue} (RGB $[0, 0, 1]$), and \textit{turquoise} (RBG $[0, 1, 0.85]$). Importantly, these pairs of values gradually decrease in distance in the RGB space. For the remaining concepts, all three tasks had the same sub-selected concept values: $shape \in [0, ..., 3], \ scale \in [0, ..., 2], \ range \in [0, 2, 4, 6], \ x\_pos \in [0,2,4,...,16], \ y\_pos \in [0,2,4,...,16]$. Thus, this experiment introduces tasks in which the loudness of the colour concept is gradually reduced.

\textbf{shapes3d} We define three tasks with the following sub-selected values:

\begin{itemize}
    \item{\textbf{Task 1}: $floor\_hue \in [0,2,...,8], \ wall\_hue \in [0,2,...,8], \ object\_hue \in [0,2,...,8], \ scale \in [0,2,...,6], \ shape \in [0,...,3], \ orientation \in [0,...,15]$}
    
    \item{\textbf{Task 2}:$floor\_hue \in [0,...,10], \ wall\_hue \in [0,...,10], \ object\_hue \in [0,...,10], \ scale \in [0,...,3], \ shape \in [0,...,3], \ orientation \in [0]$} 
    
    \item{\textbf{Task 3}:$floor\_hue \in [0,...,10], \ wall\_hue \in [0,...,10], \ object\_hue \in [0,...,10], \ scale \in [4,...,7], \ shape \in [0,...,3], \ orientation \in [0]$} 
\end{itemize}

Importantly, all three tasks have the same total number of samples ($16000$ samples). Task 1 differs from Task 2 and Task 3 by having fewer values in the first five concepts, and more values in the last concept. Task 2 differs from Task 3 by using lower values for the \textit{scale} concept, instead of higher values (hence, making it quieter).

\section{Additional Results}

\subsection{Concept-to-Task Dependence} \label{sec:appendix-task-variation-figs}

\begin{figure*}[h]
    \centering
    \includegraphics[width=\textwidth]{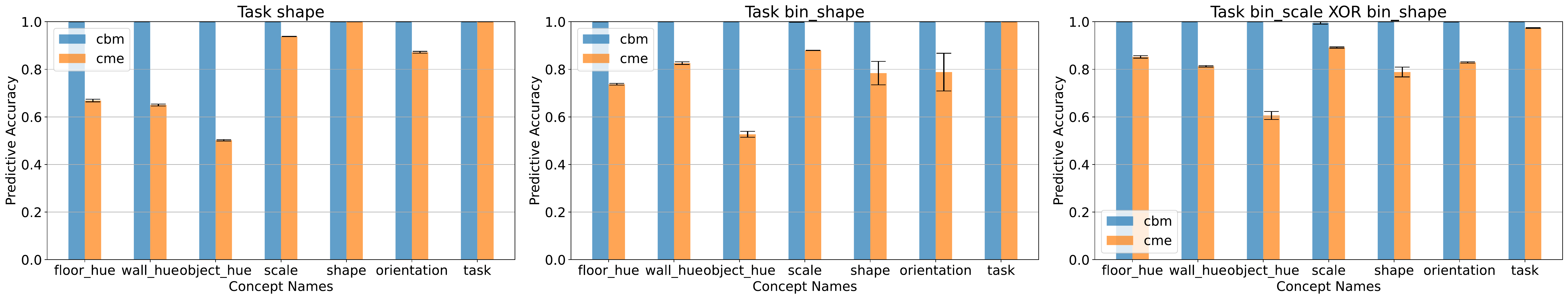}
    \caption{Comparison between the per-concept and downstream task predictive accuracy for CME and CBM across tasks of varying dependence on the concept values: full dependence on one concept (left), partial dependence on one concept (center), partial dependence on two concepts (right).}
    \label{fig:task_variance_shapes3d}
\end{figure*}

This section includes further experiments on Concept-to-Task Dependence. Figure \ref{fig:task_variance_shapes3d} shows the concept-to-task dependence results for the shapes3d dataset.

\subsection{Concept Variance Fragility} \label{sec:appendix-c-var-figs}

\begin{figure*}[h]
    \centering
    \includegraphics[width=\textwidth]{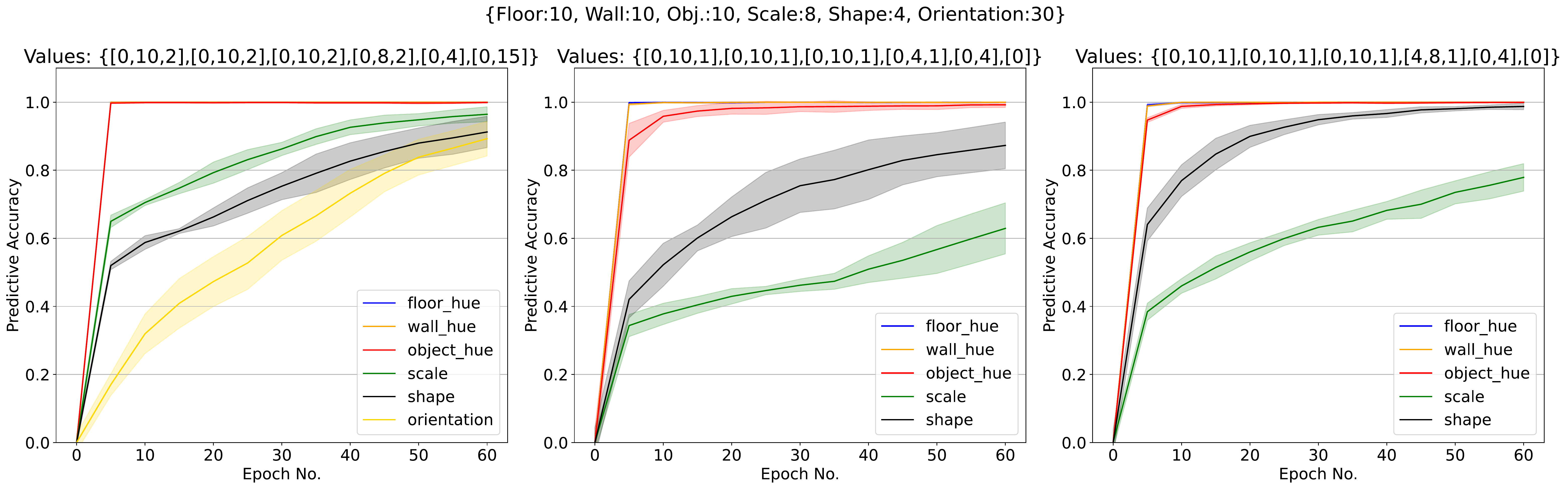}
    \caption{Comparison between different concept loudness setups for the shapes3d tasks.}
    \label{fig:concept_variance_shapes}
\end{figure*}

This section includes further experiments on Concept Variance Fragility. Figure \ref{fig:concept_variance_shapes} shows concept variance fragility for the shapes3d dataset.

\begin{figure*}[h]
    \centering
    \includegraphics[width=\textwidth]{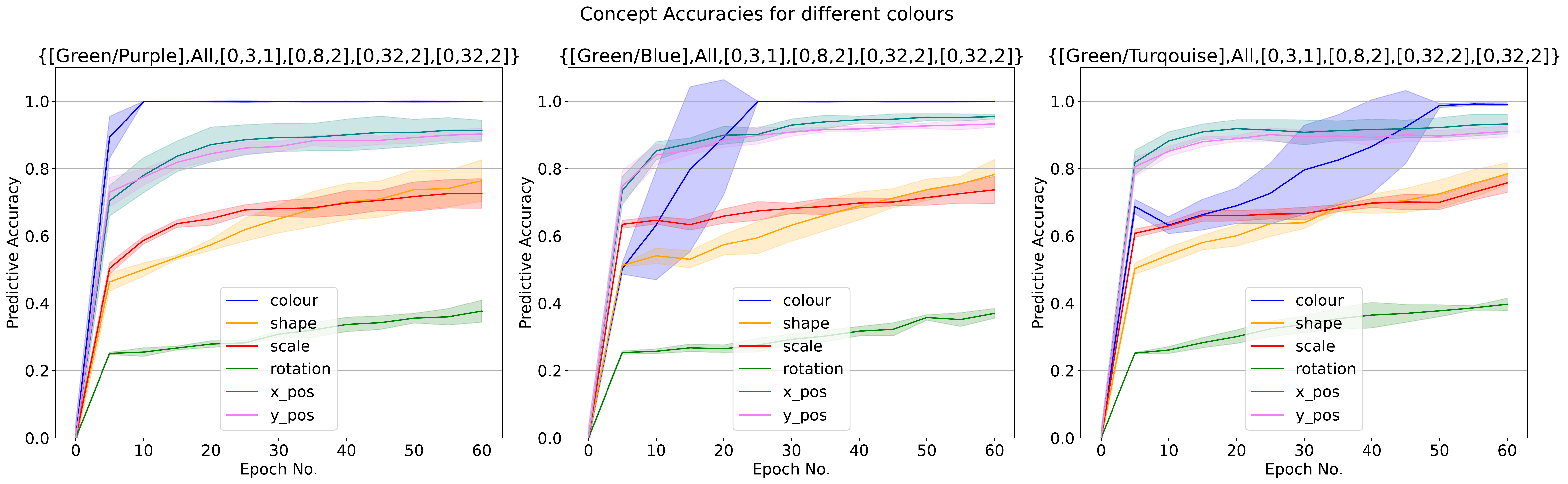}
    \caption{Comparison between different concept loudness setups for the dSprites colour tasks. Overall, the performance of the colour concept is significantly affected by its specific colour combinations.}
    \label{fig:concept_variance_clr}
\end{figure*}

Figure \ref{fig:concept_variance_clr} shows concept variance fragility for the dSprites colour experiment.

\end{document}